\def\year{2023}\relax
\documentclass[letterpaper]{article} 
\usepackage{aaai23}  
\usepackage{times}  
\usepackage{helvet}  
\usepackage{courier}  
\usepackage[hyphens]{url}  
\usepackage{graphicx} 
\def\UrlFont{\rm}  
\usepackage{natbib}  
\usepackage{caption} 
\DeclareCaptionStyle{ruled}{labelfont=normalfont,labelsep=colon,strut=off} 
\usepackage{algorithm}
\usepackage{algpseudocode}
\usepackage{paralist}
\usepackage{enumitem}
\usepackage{hhline}

\usepackage{array}
\newcommand{\PreserveBackslash}[1]{\let\temp=\\#1\let\\=\temp}
\newcolumntype{C}[1]{>{\PreserveBackslash\centering}p{#1}}
\newcolumntype{R}[1]{>{\PreserveBackslash\raggedleft}p{#1}}
\newcolumntype{L}[1]{>{\PreserveBackslash\raggedright}p{#1}}

\usepackage{newfloat}
\usepackage{listings}
\floatstyle{ruled}
\newfloat{listing}{tb}{lst}{}
\floatname{listing}{Listing}
\nocopyright
\title{Self-Adaptive Forecasting for Improved Deep Learning on Non-Stationary Time-Series}

\author{
    Sercan \"{O}. Ar{\i}k,
    Nathanael C. Yoder,
    Tomas Pfister
}
\affiliations{
    Google Cloud AI Sunnyvale, CA\\
    soarik@google.com, nyoder@google.com, tpfister@google.com
}

\usepackage{bibentry}

\begin{document}

\maketitle

\begin{abstract}
Real-world time-series datasets often violate the assumptions of standard supervised learning -- their distributions evolve over time, rendering conventional training and model selection procedures suboptimal. 
In this paper, we propose a novel method, Self-Adaptive Forecasting (SAF). SAF modifies the training of time-series forecasting models to improve their performance on forecasting tasks with such non-stationary time-series data.
SAF integrates a self-adaptation stage prior to forecasting based on `backcasting', i.e. predicting masked inputs backward in time. 
This is a form of test-time training that uses a self-supervised learning problem on test samples to train the model before performing the prediction task.
In this way, our method enables efficient adaptation of encoded representations to evolving distributions, leading to superior generalization. 
SAF can be integrated with any canonical encoder-decoder based time-series architecture, including recurrent- or attention-based ones.
On synthetic and real-world datasets in domains where time-series data are known to be notoriously non-stationary, such as healthcare and finance, we demonstrate a significant benefit of SAF in improving forecasting accuracy.
\end{abstract}
\section{Introduction}
Time-series forecasting plays a crucial role in many domains, including in finance for prediction of stock prices, in healthcare for prediction of a patient's heart rate, in retail for prediction of a product's sales and in environmental sciences for prediction of precipitation. 
Traditional approaches for time-series forecasting include statistical models, e.g. ARIMA and its extensions \citep{arima1, arima2}, exponential smoothing based methods \citep{gardner1985exponential}, and structural time-series models \citep{harvey1990forecasting}. These are recently being outperformed and replaced by high-capacity deep neural networks (DNNs), such as DeepAR \citep{deepar}, MQRNN \citep{MQRNN}, N-BEATS \citep{oreshkin2019nbeats}, Informer \citep{zhou2021informer} and TFT \citep{lim2019temporal}.

\begin{figure}[!htbp]
\centering
\includegraphics[width=0.44\textwidth]{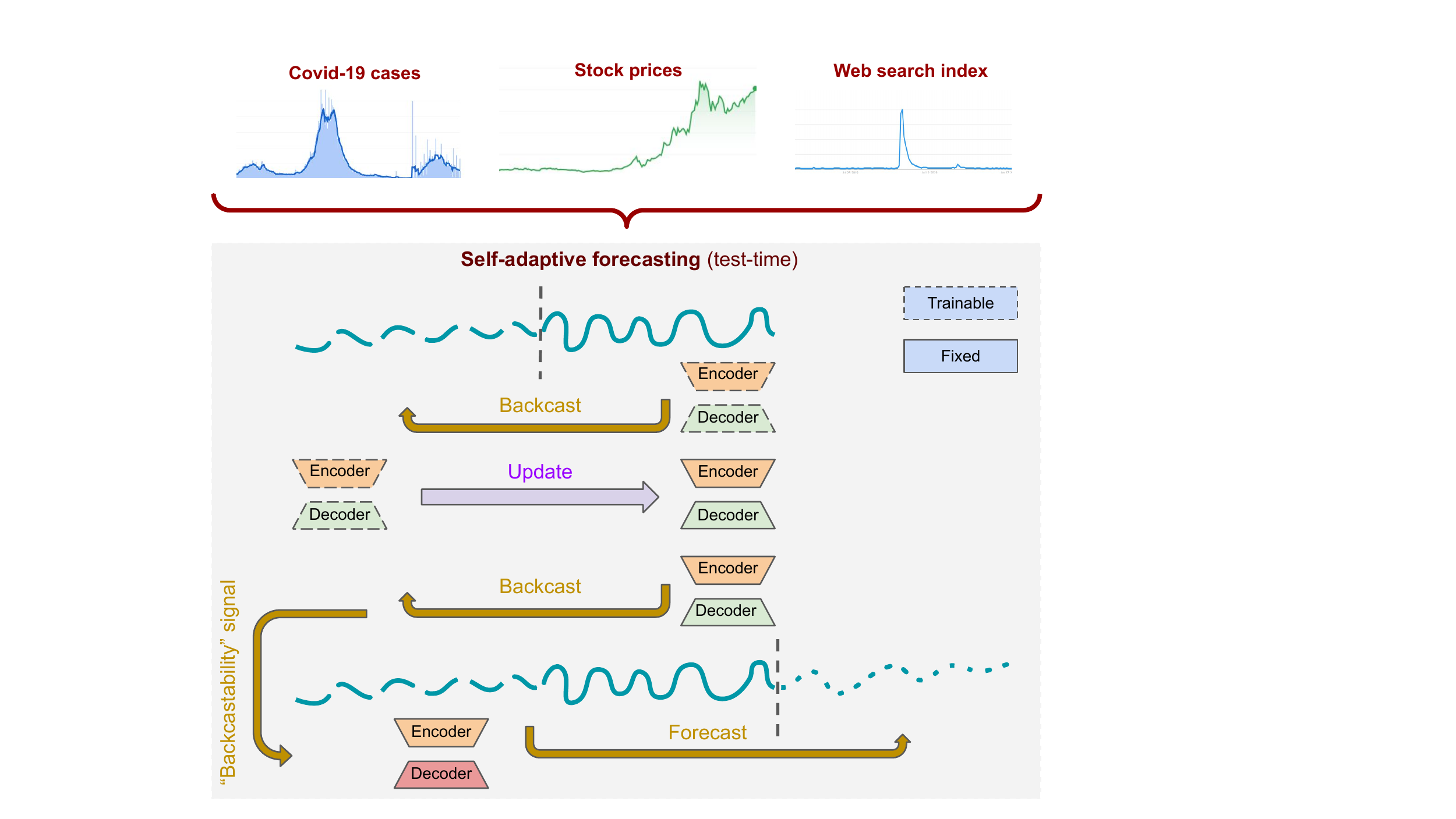}
\caption{Real-world time-series data from healthcare, finance and the web are often non-stationary, i.e. their distributions drift over time, and accurate forecasting becomes challenging. 
We propose Self-Adaptive Forecasting (SAF), which employs backcasting as a form of test-time training to update the encoders and decoders based on self-supervision and a `backcastability' error signal. 
SAF is composed of an encoder (in orange), as well as a backcast decoder (green) and a forecast decoder (red).
By adapting the model at test time according to information learned from the self-supervised loss, a test sample-specific forecasting model is learned, adapting to non-stationary conditions effectively.
}
\label{fig:motivating_examples}
\end{figure}

A prominent challenge for time-series forecasting is that many real-world time-series datasets are inherently non-stationary \cite{Kuznetsov2020}, i.e. their distributions drift over time (see Fig. \ref{fig:motivating_examples}). 
Standard supervised learning assumes that training, validation and test datasets come from the same distribution, and each sample is independently and identically distributed \citep{goodfellow2016deep}.
For non-stationary time-series data, however, these assumptions can be severely violated \citep{NIPS2015_41f1f191}. 
Consequently, the standard way of applying supervised training, used in most DNN-based time-series methods, is often suboptimal for non-stationary time-series data, yielding significant mismatch between training-validation and validation-test performances, and being detrimental for accuracy and robustness \citep{Montero2020,Kuznetsov2020}.

If data drifts are completely unpredictable, there wouldn't be much that could be done beyond improving the general robustness, particularly since DNNs are particularly sensitive to out-of-distribution data \citep{ood1, ood2}. 
Yet, for most real-world time-series data, these drifts occur with `somewhat predictable' patterns. 
For example, in financial markets, there has been a gradual increase in the amount of transactions made by trading algorithms over years \citep{multi_asset_risk}. 
While simple machine learning models might have been able to forecast the prices of financial assets a few decades ago, as these signals are exploited by such trading algorithms, the algorithms themselves change the output distributions and cause the models become less accurate. 
Although there are many other unpredictable dynamics in financial markets, this rise of algorithmic trading is one suggesting that one should not treat 1980s vs. 2010s in the same way. 
Retail is another domain with evolving dynamics, such as fashion demand cycles \citep{retail}. 
Non-stationarity is also common in healthcare, with trends like increasing obesity shaping the demand for related drugs or treatments \citep{obesity}. 
If there are time-varying features available that represent the factors causing the drift (such as fashion trend or public health indicators), they can be used as exogenous covariates input to the models \cite{lim2019temporal} so that the model can learn accurate conditional distributions based on them. 
However, in practice readily-available covariates cannot fully capture all data drifts. 
Therefore, it is strongly desired to be able to implicitly learn such changes and relate to exogenous signals whenever possible. 

In this paper, we propose a canonical framework to learn generating accurate forecasts for non-stationary time-series data. 
Self-Adaptive Forecasting (SAF) is an architecture-agnostic method for forecasting non-stationary time-series, that allows forecasting models to better adapt to current dynamics (see Fig. \ref{fig:motivating_examples}).
SAF is based on test-time training, which is motivated by the realization that since test samples give a hint about the underlying test distribution, it makes sense to adapt the model at test time according information learned from the test samples.
Self-supervision can be used to obtain a test-sample specific model that is used to make the final prediction. 
In SAF, encoded representations are updated with a backcasting self-supervised objective, defines as the task of predicting the masked portion of the time-varying input features backwards in time. 
With this adaptation step, the representations are encouraged to consider the input joint distribution. 
By better reflecting the current state of the varying distributions to condition the generated forecasts (and by having a matching end-to-end training procedure), SAF significantly improves the forecasting accuracy, especially so for datasets with varying dynamics.
Our contributions can be summarized as:
\begin{compactenum}
\item We propose a novel method, SAF, for improved performance non-stationary time-series forecasting.
\item To our knowledge, SAF is the first application of test-time training concept for time-series forecasting.
\item SAF can be integrated into any encoder-decoder based DNN architecture, and we demonstrate its efficacy with recurrent- and attention-based architectures. 
\item Specifically on synthetic and real-world data with severe non-stationarity, we demonstrate significant performance improvements with SAF across various datasets with different characteristics.
\end{compactenum}

\section{Related work}

\textbf{Forecasting non-stationary time-series:}
Various statistical estimation and machine learning methods have been proposed for non-stationary time-series forecasting, often based on integrating an adaptive component in traditional supervised learning formulations. 
One approach is modifying the estimation procedure for autoregressive processes with time-varying coefficients \citep{Dahlhaus}.
\citep{BRAHIMBELHOUARI2004705} studies Gaussian process models using a non-stationary covariance function.
\citep{recursive_estimation_nonstationary} proposes a spectral decomposition procedure, based on the exploitation of recursive smoothing algorithms, for self-adaptive implementation of state-space forecasting and seasonality adjustments. 
\citep{dynamic_svm} modifies support vector machines using an exponentially increasing regularization constant and an exponentially decreasing tube size, to deal with structural changes in the data. 
\citep{NIPS2015_41f1f191, Kuznetsov2020} proposes a convex learning objective with learning guarantees for the general case of non-stationary non-mixing processes. 
\citep{app9071345} studies wavelet decomposition-based approaches, which are flexible in modeling local spectral and temporal information that allows capturing short duration, high frequency, longer duration, and lower frequency information simultaneously.
Overall, these previous works are model-dependent and are not straightforward to integrate with deep learning. 

\noindent \textbf{Deep learning under distribution shift:}
Numerous works demonstrated that DNNs can severely suffer as the test distributions deviate from training distributions \citep{ovadia2019trust,Mandoline,wilds}. 
Various methods were proposed to mitigate such performance degradation, particularly for visual data.
Domain-adversarial deep learning \citep{ganin2016domainadversarial} is one common approach for visual data that encourages the emergence of features that are discriminative for the main source domain task and indiscriminate with respect to the shift between the domains.
\citep{Contrastive_adaptation} adapts this by optimizing a metric which explicitly models the intra-class domain
discrepancy and the inter-class domain discrepancy. 
\citep{Fang_rethinking} proposes dynamic importance weighting which contains estimation of the test-over-training density ratio and fitting the classifier from weighted training data, showing promising results on image classification tasks. 
In general, the methods in this space are not designed to utilize the time component and are often not implemented in online fashion (i.e. they expect the test distribution at test time). Therefore, such methods are not commonly used for time-series forecasting.
To significantly improve non-stationary time-series forecasting with DNNs, a method that explicitly uses the time component along distribution shift is highly desired.

\noindent \textbf{Test-time training:}
Test-time training is motivated by the realization that test samples can be used for training for better test accuracy, as a form of one sample learning. 
Instead of anticipating the test distribution at training time (as in domain adaptation), the idea is to instead adapt the model at test time, according information learned from the test samples.
In practice, this is achieved by creating a self-supervised objective function that is optimized both at training and testing time.
At training time, the self-supervised and supervised losses should be used as well, mimicking the testing time scenario. 
At testing time, the self-supervised loss is optimized on test samples using gradient descent, and the test sample-specific model is used to make the final prediction. 
\cite{ttt} demonstrated the benefits for improving image classification accuracy particularly under distribution shifts, using rotation prediction as the self-supervised objective. Beyond multi-task learning, \citep{bartler2021mt3} proposes to modify training with gradient-descent based meta learning \citep{maml} to better mimic the test-time adaptation task. 
Our proposed training methods follow the same rationale for applying gradient-descent based meta learning-based training.
The test-time training idea has also been adapted to reinforcement learning via self-supervised policy adaptation \citep{hansen2021selfsupervised}. 
To our knowledge, our work is the first application of test-time training concept to time-series data and for forecasting, utilizing the time-component for the self-supervised objective and adaptation at test time.

\begin{figure}[!htbp]
\centering
\includegraphics[width=0.4\textwidth]{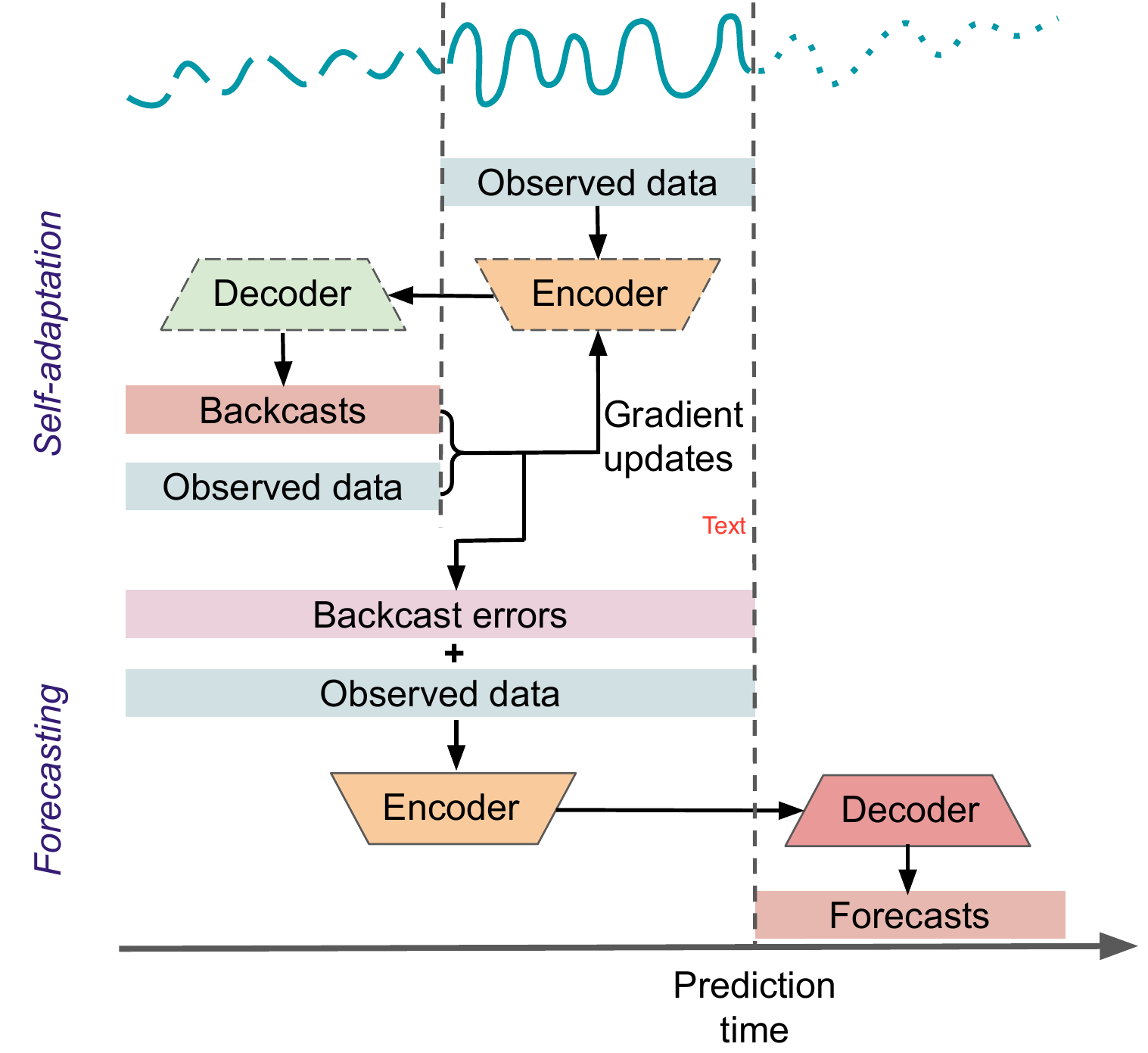}
\caption{Self-Adaptive Forecasting (SAF): improve deep learning based time-series forecasting by learning to better adapt to current dynamics using self-supervision.
During the test-time self-adaptation stage, a masked portion of the input is backcasted (predicting the past), and the gradients of the backcasting loss are used to adapt the encoder (the dashed box denotes it being trainable initially and then the solid box denotes it being used in feed-forward way). 
During the forecasting stage, this test window-specific adapted encoder is used together with the backcast errors (which provide a signal of `backcastability') to perform the forecast.}
\label{fig:main_figure}
\end{figure}

\section{Principles behind self-adaptive forecasting}

We propose to modify deep learning based forecasting to address the bottlenecks of non-stationarity time-series data, via minimizing the impact of distribution mismatch by utilizing the predictable patterns along time. 
Fig.~\ref{fig:main_figure} overviews the general idea of the proposed Self-Adaptive Forecasting (SAF). The desiderata of SAF are described below: 

\textbf{Conditioning forecasting on the current state}: 
We first start with the motivation that the forecasting should be conditioned not only on input covariates, but also on an abstract notion of `state', which should reflect the data distribution around the prediction time. 
These `states' should be powerful representations of the time-series data in the encoding window, and the distribution they reflect should be globally meaningful. 
For example, if a very similar regime is observed in the past, a similar state should be retrieved with the value corresponding to that past timestep. 
Such `states' should be learned from data around the prediction time, and directly affect the forecasts.
One idea for obtaining such states is to use explicit learnable representations to condition the models on, and fit them separately.
Instead, we propose to directly reflect the states within encoder weights implicitly, dynamically updating the encoder weights so that they efficiently represent the given window.  

\textbf{Test-time training for determining the current state}:
Determining the states should be done in a `self-supervised' way as the information for future timesteps, the target, would be unknown at test time. 
Accordingly, we propose to modify the forecasting operation based on the updates from self-supervised learning at test time. 
Our modification is based on applying a gradient-descent update with the self-supervised objective. 
In contrast to learning methods that modify the loss function only during training, such as multi-task learning or autoencoding, test-time training has a loss function for model modification at test time (see Fig. \ref{fig:motivating_examples}), which helps to adapt different conditions effectively. 

\textbf{Judicious self-supervised learning}: 
It is critical that the self-supervised learning task should be meaningful for the time-series data so that the model can utilize the most sentient representation related to the forecasting task. 
One path is to propose a self-supervised learning task with domain knowledge (analogous to rotation prediction for object-centric images \cite{ttt}) -- for example, pitch prediction could be a meaningful task for speech time-series data \citep{selfsupverised_pitch}. 
Yet, to apply our method to wide range of datasets, we propose a canonical self-supervised learning task for forecasting: `backcasting', where the task is to predict the inputs (`back in time') in a masked portion of the window. Our motivation for backcasting is based on the intuition that as the past-future relationship evolves along time, the predictability of the relationship at different consecutive timesteps should reflect the evolution. An unexpected event that deteriorates the validity of the learned past-future relationship should show up in the achievable backcasting, as well as achievable forecasting.
The backcasting loss metric can be similar to the forecasting loss metric, such as L1/L2 distance based regression losses.
In contrast to the core forecasting task at which one often only needs to predict one target covariate (e.g. sales for retail demand forecasting), the self-supervised backcasting cast can benefit from being applied to all input covariates (e.g. number of customers, economical indices, weather etc.) to better learn varying distributions by jointly modeling all covariates from training data. 

\textbf{Self-supervised error signal}: 
If there is a large amount of change in the distribution and the current state is very unpredictable, the forecaster component should utilize it and adjust accordingly. 
For example, the forecaster can output more typical values rather than an outlier in the presence of unpredictable scenarios. 
To that end, we propose to input the error signal from the self-supervised learning adaptation. We keep the use of this error signal as optional -- whether to use it or not is a hyperparameter and picked based on the validation metric.

In the next section, we describe our specific implementation that utilizes conventional building blocks and optimization methods built on these principles.

\begin{algorithm*}[!ht]
\caption{Inference for proposed Self-Adaptive Forecasting (SAF).}
\begin{algorithmic}[1]
\Require{Input instance $x$, encoder $e(\cdot; \mathbf{\Omega}_e)$, backcast decoder $d^{(b)}(\cdot; \mathbf{\Omega}_{d^{(b)}})$, forecast decoder $d^{(f)}(\cdot; \mathbf{\Omega}_{d^{(f)}})$, prediction time $t$, mask length $n$, and input window length $m$.}
    \State $\mathbf{\Theta} = \textit{Tile}(x[t{-}m{+}n{+}1])$ \Comment{Tile the last element of the unmasked window.} 
    \State $\mathbf{r} = e( [\mathbf{\Theta}, x[t{-}m{+}n{+}1:t]] \cup \mathbf{0}; \mathbf{\Omega}_e)$  \Comment{Obtain encoded representation for the unmasked input window.} 
    \State $b = d^{(b)}(\mathbf{r}; \mathbf{\Omega}_{d^{(b)}})$  \Comment{Obtain backcasts for the masked window.} 
    \State $\mathbf{\Omega}^{(m)}_e \leftarrow \mathbf{\Omega}_e - \alpha \cdot \nabla_{\mathbf{\Omega}_e} L(b, x[t{-}m{+}1:t])$  \Comment{Self-adaptation: Update the encoder with backcast loss.} 
    \State $\mathbf{\Omega}^{(m)}_{d^{(b)}} \leftarrow \mathbf{\Omega}_{d^{(b)}} {-} \alpha \cdot \nabla_{\mathbf{\Omega}_{d^{(b)}}} L(b, x[t{-}m{+}1:t])$  \Comment{Self-adaptation: Update the backcast decoder with backcast loss.} 
    \State $b = d^{(b)}(\mathbf{r}; \mathbf{\Omega}^{(m)}_{d^{(b)}})$  \Comment{Obtain backcasts for the masked window.} 
    \State $\mathbf{e} = x[t{-}m{+}1:t] - b$  \Comment{Obtain errors for the backcasts.} 
    \State $\mathbf{r} = e(x[t{-}m{+}1:t] \cup \mathbf{e}; \mathbf{\Omega}^{(m)}_e)$  \Comment{Obtain encoded representation for the input window combined with the errors.} 
    \State $\hat{y} = d^{(f)}(\mathbf{r}; \mathbf{\Omega}^{(m)}_{d^{(f)}})$ \Comment{Forecast.} 
\end{algorithmic}
\end{algorithm*}

\section{Implementation}

Let's consider the multi-horizon forecasting task of predicting $y[t+1:t+h]$ from the observed inputs $x[t-m+1:t]$, where $h$ is the forecasting horizon and $m$ is the input window length.\footnote{$[\cdot, \cdot]$ assumes inclusion of the boundary values.}
We focus on a canonical multi-horizon forecasting architecture with an encoder $e(\cdot; \mathbf{\Omega}_e)$, and forecast decoder $d^{(f)}(\cdot; \mathbf{\Omega}_{d^{(f)}})$. 
In addition to these two modules, SAF introduces a backcast decoder $d^{(b)}(\cdot; \mathbf{\Omega}_{d^{(b)}})$. 
The architecture of the backcast decoder can be similar to a forecast decoder as it maps the encoded representations at the corresponding timesteps to backcasts, similar to how the forecasting decoder maps the encoded representations at the corresponding timesteps to forecasts.
For example, LSTM sequence-to-sequence models \citep{sutskever2014sequence} may employ separate LSTMs as the encoder, forecasting decoder and backcasting decoders, and the encoded representation may be used as an input to the decoders via the initial state. 
In contrast, attention-based models such as TFT \citep{lim2019temporal} may contain a self-attention module in the encoder (along with temporal representation learning), and decoders may be multi-layer perceptrons (MLPs) that map the encoded representations of the timesteps to the backcasts or forecasts. 
There is no restriction in the encoder and decoder architectures -- they may contain other types of layers, such as convolutional or relational. 

We start the explanation of the SAF implementation with the inference procedure, and as we describe later, the training procedure is designed to match the desired inference capability. Overall, Algorithm 1 overviews how the forecasting is performed with SAF.

To adapt to the evolving distributions, we utilize encoded representations $\mathbf{r}$ to represent the current `states' of the time-series. 
$\mathbf{r}$ is obtained based on the proposed self-supervised learning task of backcasting of the window, for which we propose masking a portion of the input in the encoded window, $[t-m+1:t-m+n]$, and predicting the masked portion with the unmasked portion, $[t-m+n+1:t]$. 
To reuse the encoder for the same length, the elements from the masked portion of the input $x[t-m+1:t-m+n]$ are replaced with the first observed element of the unmasked portion $x[t-m+n+1]$, as opposed to naively inputting an arbitrary constant that may exacerbate covariate shifts and thus training issues \cite{covariate_shift}. (Alg. 1, Step 1 - here $Tile$ operation corresponds to constructing the vector with repeated elements). 
In our experimental results, we simply mask half of the window, i.e. $n=m/2$, as it is observed as a reasonable choice across various datasets. $n/m$ ratio can also be treated as a hyperparameter.

The encoded representation $\mathbf{r}$ is input to the backcasting decoder, which produces the backcasts $\mathbf{b}$ to be used in the loss $L(\mathbf{b}, x[t-m+1:t])$. 
The loss is used to supervise the encoder and backcast decoder based on a single-step gradient descent update (Alg. 1, Steps 4 and 5) -- i.e. they are adapted to more accurately backcast the input.
We note that multiple gradient descent steps can also be used for better optimization, but we limit to a single step to yield low computational complexity overhead.
Each inference run modifies the pre-trained encoder $e(\cdot; \mathbf{\Omega}^{(m)}_e)$ (Alg. 1, Step 4) and backcast decoder $d^{(b)}(\cdot; \mathbf{\Omega}^{(m)}_{d^{(b)}})$ (Alg. 1, Step 5) with the gradient descent updates. 
To ensure that the original pre-trained model is not altered between different inference runs and the model outputs do not change for multiple applications of the same model, we apply them to a different copy of weights (denoted with the $^{(m)}$ superscript). 

After the self-adaptation update, the encoded representations $\mathbf{r}$ are re-obtained with the updated encoder $e(\cdot; \mathbf{\Omega}^{(m)}_e)$ (Alg. 1, Step 6). 
In addition, we also re-obtain the backcasts and estimate the backcasting error $\mathbf{e}$ (Alg. 1, Steps 6 and 7) with the goal of providing useful signal on the difficulty of prediction of a particular timestep, which could be useful in indicating the level of non-stationarity, as explained above. 
We propose to use this backcasting error~$\mathbf{e}$ in the encoder as an additional time-varying feature (appended to as a new feature) along with the input in the entire time window. 
The forecasts are obtained from the encoded representation of the entire input and the error signal, $\mathbf{r} = e(x[t-m+1:t] \cup \mathbf{e})$. 
If used, error values at different timesteps are combined as additional features wherever they exist to be fed into the encoder architecture; otherwise $\mathbf{0}$ is used.
Note that we do not have the error for backcasting to fed into the encoder -- instead we input $\mathbf{0}$ (Alg. 1, Step 2).
Finally, the forecasting decoder generates the multi-horizon forecasts as the outputs (Alg. 1, Step 9) using the updated encoded representations. 

For training, matching the inference procedure is essential to optimize for the actual test-time functionality given in Algorithm 1. Our objective is to learn the encoder $e(\cdot; \mathbf{\Omega}_e)$, backcast decoder $d^{(b)}(\cdot; \mathbf{\Omega}_{d^{(b)}})$, forecast decoder $d^{(f)}(\cdot; \mathbf{\Omega}_{d^{(f)}})$ from the training split of the time-series data. For update of the encoder and forecasting decoder, supervision from the forecasting loss is employed.
We note that the proposed training method is applicable even with batches containing different time-series entities and prediction timesteps, as they have an encoding window of certain fixed duration and they can be combined in the same training batch in a computational efficient way. 
The full training procedure is presented in the Appendix.

\begin{table*}[!htbp]
\caption{Test MSE on the 4 synthetic autoregressive processes datasets, averaged over 11 different durations (with ${\pm}$ standard deviations for the aggregate statistics over different time-series durations). For all cases, the hyperparameters are picked based on the validation MSE.
A significant improvement is observed on the two non-stationarity datasets (AR1\& AR3) with up to $\sim$10\% reduction for average MSE.}
\resizebox{\textwidth}{!}{
\begin{tabular}{|c|C{3.5cm}|C{3.5cm}|C{3.5cm}|C{3.5cm}|}
 \cline{1-5}
 & \textbf{AR1: Abrupt change} & \textbf{AR2: Smooth drift} & \textbf{AR3: Random state change} & \textbf{AR4: Stationary}  \\ \hhline{|=|=|=|=|=|}
LSTM Seq2Seq & .002601 ${\pm}$ .000822 & .001370 ${\pm}$ .000750 & .002061 ${\pm}$ .001338 & \textbf{.001167 ${\pm}$ .000159} \\ \cline{1-5}
LSTM Seq2Seq - SAF & \textbf{.002550 ${\pm}$ .000718 (-1.97\%)}  & \textbf{.001319 ${\pm}$  .000692 (-3.73\%)} & \textbf{.001859 ${\pm}$ .000616 (-9.81\%)} & .001185 ${\pm}$ .000181 (1.54\%) \\ \hhline{|=|=|=|=|=|}
TFT  & .002744 ${\pm}$ .000821 & \textbf{.001416 ${\pm}$ .000841} & .001823 ${\pm}$ .000815 & .001206 ${\pm}$ .000180\\ \cline{1-5}
TFT - SAF & \textbf{.002540 ${\pm}$ .000729 (-7.44\%)} & .001426 ${\pm}$ .000881 (0.70\%) & \textbf{.001659 ${\pm}$ .000525 (-9.00\%)} & \textbf{.001164 ${\pm}$ .000171 (-3.49\%)} \\ \cline{1-5}
\cline{1-5}
\end{tabular}}
\label{fig:syn_results}
\end{table*}

\section{Forecasting performance}

We perform experiments on various non-stationary time-series datasets to evaluate the efficacy of the proposed SAF. 
Our method is model architecture agnostic, and we demonstrate its potential with an LSTM sequence-to-sequence model and an attention-based model, TFT. 
For the LSTM sequence-to-sequence model, the backcast decoder $d^{(b)}(\cdot; \mathbf{\Omega}_{d^{(b)}})$ is based on another LSTM. 
Static features are mapped with an MLP, and the static representations are either added or concatenated to the encoded representations coming from the LSTM (the choice is treated as a hyperparameter, see the Appendix). 
For TFT, it is based on a dense layer that maps the prediction from attention outputs, in the same way that the dense layer in standard TFT maps the attention outputs to forecasts. 
Some datasets contain multiple entities (i.e. multiple target time-series). 
For all experiments, we consider datasets with training-validation-test split over time.
Model selection is based on the validation performance, and we evaluate the test performance. 
The overall goal is to highlight how SAF can improve conventional architectures and training approaches.
Please see the Appendix for further details datasets and training.

\begin{table*}[!htbp]
\caption{Properties of the real-world datasets used in experiments. Details are provided in Appendix.}
\centering
\begin{tabular}{|c|c|c|c|c|c|}
\cline{1-6}
 & \textbf{Covid-19} & \textbf{M5} & \textbf{Electricity} & \textbf{Crypto}  & \textbf{SP500}  \\ \hhline{|=|=|=|=|=|=|}
Sampling frequency & Daily  & Daily & Hourly & Minutely & Daily \\ \cline{1-6}
Number of entities & 56  & 200 & 369 & 14 & 75 \\ \cline{1-6}
Forecasting horizon & 28  & 28 & 24 & 1 & 1 \\ \cline{1-6}
Number of time-varying features & 20 & 13 & 9 & 8 & 6 \\ \cline{1-6}
Number of static features & 15 & 5 & 2 & 3 & 1 \\ \cline{1-6}
Number of train timesteps & 283  & 1707 & 5471 & 1.9M & 2074 \\ \cline{1-6}
Number of validation/test timesteps & 14  & 14 & 168 & 2160 & 30 \\ 
\cline{1-6}
\end{tabular}
\label{table:datasets}
\end{table*}

\subsection{Synthetic datasets}

We consider 4 autoregressive random processes from \citep{Kuznetsov2020} with additive Gaussian noise $\epsilon_t$ (with a mean of 0 and standard deviation of .03):

\noindent \textbf{AR1:} Includes abrupt changes in the data generating mechanism: $y[t] = \alpha[t] \cdot y[t-1] - \epsilon_t$, where $\alpha[t]=-0.9$ for $t \in [1000, 2000]$ and 0.9 otherwise.

\noindent \textbf{AR2:} Has parameters for the data generating process smoothly drifting: $y[t] = \alpha[t] \cdot y[t-1] - \epsilon_t$, where $\alpha[t]=1-(t/1500)$.

\noindent \textbf{AR3:} Has parameter changes occurring at random times: $y[t] = \alpha[t] \cdot y[t-1] - \epsilon_t$, where $\alpha[t]$ is either -0.5 or 0.9 based on the stochastic process, which after spending $\tau$ last time steps, at the next time step will stay in the same one with probability $(0.99995)^\tau$ and will move to the different state with probability $1-(0.99995)^\tau$.

\noindent \textbf{AR4:} With stationarity: $y[t] = -0.5 \cdot y[t-1] - \epsilon_t$.

We consider 11 different total dataset durations in the range [1000, 3000] with increments of 200, and a forecasting horizon of $h=5$. 
We use validation and test durations of 100 for each case and the rest of the data is used for training. 
Mean squared error (MSE) loss is used for training and evaluation (given the additive noise has Gaussian distribution).

Table \ref{fig:syn_results} shows the overall results on these 4 datasets. 
We observe the benefit of SAF with both LSTM Seq2Seq and TFT models, especially on the two non-stationarity datasets (AR1 \& AR3) with abrupt or random state shifts -- SAF yields up to $\sim$10\% reduction for average MSE. On AR2, with smooth drift, the benefit is smaller for LSTM Seq2Seq, and we observe slight underperformance with TFT. 
On AR4, the stationary process, SAF gives slightly worse results than the baseline, which we attribute to distracting the model capacity with the self-supervised task when the supervised task itself is too simple for model to learn across available data.

\subsection{Real-world datasets}

We consider real-world time-series datasets from Healthcare, Retail, Energy and Finance domains. 
We intentionally pick datasets with different characteristics (summarized in Table \ref{table:datasets}) to demonstrate the applicability of SAF across different regimes. 
The Appendix contains additional details about the datasets.

\noindent \textbf{Covid-19:}
We consider the task of forecasting the state-level Covid-19 deaths for the next 28 days from various time-varying and static covariates as in \cite{covid_forecasting}. 

\noindent \textbf{M5:}
We use the M5 competition dataset\footnote{\url{https://www.kaggle.com/c/m5-forecasting-accuracy/}} for the task of forecasting the unit sales of retail goods.

\noindent \textbf{Electricity:}
We consider the task of forecasting electricity consumption of different customers on the UCI Electricity Load Diagrams dataset \cite{NIPS2016_85422afb}.

\noindent \textbf{Crypto:}
We use the G-Research Crypto Forecasting dataset\footnote{\url{https://www.kaggle.com/c/g-research-crypto-forecasting/}} for short term returns in 14 popular cryptocurrencies, based on the prediction targets of the competition. 

\noindent \textbf{SP500:}
We consider the task of forecasting the daily returns at open for 75 equities from SP500 index (which have data since 01-01-2012) using Kaggle's `Stock Market Data'\footnote{\url{https://www.kaggle.com/paultimothymooney/stock-market-data}}.

\begin{table*}[!htbp]
\caption{Test set results on the real-world datasets. The metrics are MAE except M5 and Crypto, for which we use MSE (following the metric choices of related literature for these problems). For all cases, the hyperparameters are picked based on the validation metric. SAF provides consistent improvements with both LSTM Seq2Seq and TFT across all datasets.}
\centering
\begin{tabular}{|c|c|c|c|c|c|}
\cline{1-6}
 & \textbf{Covid-19} & \textbf{M5} & \textbf{Electricity} & \textbf{Crypto}  & \textbf{SP500}  \\ \hhline{|=|=|=|=|=|=|}
LSTM Seq2Seq, baseline & 1357.1  & 5.741 & 45.95 & 3.592  & 0.6836 \\ \cline{1-6}
LSTM Seq2Seq, SAF & 793.5 (\textbf{-41.6\%}) & 5.474 (\textbf{-4.6\%}) &  44.53 (\textbf{-3.6\%}) & 3.459 (\textbf{-3.8\%}) & 0.6656 (\textbf{-2.7\%}) \\ \hhline{|=|=|=|=|=|=|}
TFT, baseline  & 1394.7 &  5.817 & 43.10 & 4.249 & 0.6693 \\ \cline{1-6}
TFT, SAF & 1117.9 (\textbf{-19.9\%}) & 5.609 (\textbf{ -3.6\%})   & 40.89  (\textbf{ -5.4\%}) & 3.459 (\textbf{-18.6\%}) & 0.6689 (\textbf{-0.1\%}) \\ \cline{1-6}
\cline{1-6}
\end{tabular}
\label{fig:real_results}
\end{table*}

Table \ref{fig:real_results} presents the test set accuracy performance on the above real-world datasets for both LSTM Seq2Seq and TFT-based architectures.
Overall, SAF provides consistent improvements with both LSTM Seq2Seq and TFT architectures across all cases, with some variance in improvements depending on the dataset. 
The largest accuracy improvement is on Covid-19, with ${>}40\%$ error reduction for LSTM Seq2Seq, and ${>}19\%$ for TFT. 
Covid-19 dataset is known to possess highly non-stationarity dynamics as noted by different pandemic studies \citep{covid_forecasting} (given the rapid changes in disease dynamics caused by non-pharmaceutical interventions like mask mandates or school closures, vaccination and virus mutations).
Next section analyzes forecast improvement scenarios further for Covid-19.
Following, the biggest improvement is observed for the Crypto dataset, which is also known to possess high non-stationarity in its dynamics \citep{matic2021hedging}. The dataset on which we show the smallest improvement, SP500, tends to yield similar results with different models and hyperparameters, which we attribute to limited predictability of the target given the features as the loss also cannot decrease much. On M5 and Electricity datasets, two popularly used academic benchmarks, SAF yields ${>}3\%$ improvements in all cases for both architectures.

\begin{figure*}[!htbp]
\centering
\includegraphics[width=0.33\textwidth]{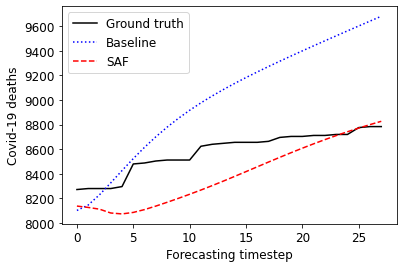}
\includegraphics[width=0.33\textwidth]{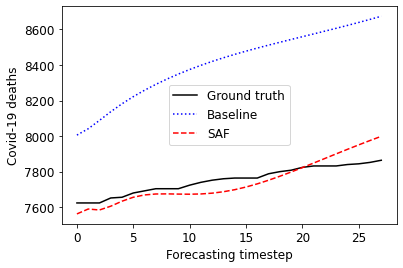}
\includegraphics[width=0.33\textwidth]{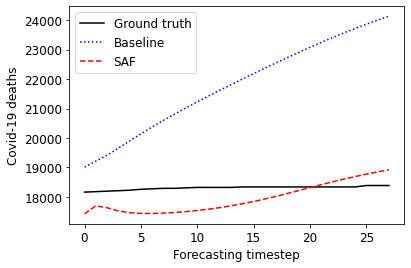}
\caption{Ground truth vs. forecasts with LSTM Seq2Seq on Covid-19 for 3 US states, without (blue) and with (red) SAF.}
\label{fig:example_forecasts}
\end{figure*}

\section{Performance analyses}

In this section, we shed further light on how SAF helps to improve the forecasting accuracy.

\noindent \textbf{Capturing time-series characteristics}:
Fig. \ref{fig:example_forecasts} exemplifies scenarios where the forecast accuracies are improved significantly on Covid-19. 
SAF provides benefits in learning the overall trend behavior accurately and preventing unrealistic exponential growth, a typical problem suffered by autoregressive models like LSTM for long-horizon prediction \cite{LINDEMANN2021650, 8490843}.
In addition, SAF can help learning the forecast for the first timestep more accurately, thanks to the backcasting objective, -- it discovers the pattern that Covid-19 cumulative deaths on the first prediction date should not be significantly different than the last observed cumulative value. 
Note that such knowledge is not embodied in the black-box LSTM architecture and the model needs to learn it from the limited amount of training data. 
On the other hand, the inductive bias created by SAF (as opposed to an approach with hand-crafted inductive bias creation) readily helps with this, as the backcasting task also provides supervision to learn the continuity of the cumulative quantities. 

\noindent \textbf{Number of time-varying features}:
Although the core forecasting task focuses on predicting one target covariate, the self-supervised backcasting can benefit from being applied to all input time-varying covariates. 
We observe that on the datasets with a high number of time-varying covariates (see Table \ref{table:datasets}), such as Covid-19, we tend to observe bigger benefit of SAF. 
As time-series datasets contain more and more covariates (given the corresponding data infrastructures are becoming more widely available to join many relevant time-series signals), they are positioned for superior time-series forecasting by adapting SAF.

\begin{figure}[!htbp]
\centering
\includegraphics[width=0.38\textwidth]{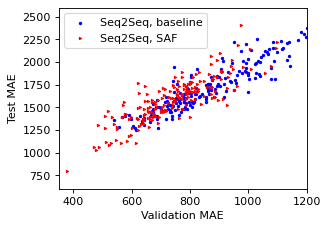}
\caption{Validation vs. test MAE for different hyperparameter trials (each marker corresponds to a different set of hyperparameter configurations) for LSTM Seq2Seq on Covid-19. 
SAF improves both the achievable validation performance and the validation and test correlation across runs.}
\label{fig:hparam_analysis}
\end{figure}

\noindent \textbf{Robust hyperparameter selection}: 
One of the benefits of SAF is reducing the mismatch between validation and test metrics, helping to select a better model. 
Fig. \ref{fig:hparam_analysis} shows the validation and test results of the individual hyperparameter trials. 
As can be observed, SAF not only gives trials with better validation error, but indeed the correlation between validation and test errors improves, especially for trials with low validation error. 
For example, the Spearman's rank correlation coefficient is 0.706 for the LSTM Seq2Seq baseline without SAF vs. 0.775 with SAF for the trials with validation MAE less than 800 (Fig. \ref{fig:hparam_analysis}). 
Improved correlation between validation and test metrics yields superior hyperparameter optimization and thus generalization, without expensive cross-validation approaches.

\noindent \textbf{Ablation studies}: 
We perform ablation studies by removing some major constituents of SAF, on Covid-19.
Table \ref{fig:ablations} demonstrates the importance of having the gradient-descent based update steps in self-adaptation stages for encoder (Alg. 1, Step 4) and decoder (Alg. 1, Step 5), which ties back to the fundamental principle of test-time adaptation based model modification to better reflect the current `state'. In addition, the self-supervised error signal (given in Alg. 1, Step 8) is observed to be quite helpful. The information captured by it prevents the model from making highly off forecasts when the backcasting is very challenging.

\begin{table}[!htbp]
\caption{Ablation studies on Covid-19 with LSTM Seq2Seq.}
\centering
\begin{tabular}{|c|c|}
\cline{1-2}
{\bf Ablation scenario} & \textbf{Test MAE}  \\ \hhline{|=|=|}
SAF without updating the decoder & 1421.1 \\ \cline{1-2}
SAF without updating the encoder & 1157.7 \\ \cline{1-2}
SAF without error signal  & 941.2 \\ \cline{1-2}
SAF & {\bf 793.5} \\ \cline{1-2}
\cline{1-2}
\end{tabular}
\label{fig:ablations}
\end{table}

\section{Conclusions}
We propose self-adaptive forecasting (SAF), a novel approach to improve deep learning performance on non-stationarity time-series data. 
SAF integrates a self-adaptation stage prior to forecasting based on `backcasting' the masked inputs, a form of test-time training. 
This enables adapting encoded representations efficiently to evolving distributions, resulting in better generalization. 
SAF can be integrated with any canonical encoder-decoder based time-series forecasting model, including recurrent neural network or attention-based architectures. 
On various synthetic and real-world non-stationary time-series datasets with distinct characteristics, we show that SAF can result in significant improvements in forecasting accuracy. 
We believe SAF can open new horizons on how to design learning methods for time-series forecasting in robust way, a major need from many real-world applications. 

\newpage
\section{Appendix}

\begin{algorithm*}[!htbp]
\caption{Training for proposed self-adaptive forecasting.}
\begin{algorithmic}[1]
\Require{Training dataset $S = \{(x, y) \}$; encoder $e(\cdot; \mathbf{\Omega}_e)$, backcast decoder $d^{(b)}(\cdot; \mathbf{\Omega}_{d^{(b)}})$, forecast decoder $d^{(f)}(\cdot; \mathbf{\Omega}_{d^{(f)}})$, mask length $n$, and input window length $m$.} 
\While{Until convergence}  
    \State $(x_B, y_B) \in S$  \Comment{Sample a batch.}
    \State $\mathbf{\Theta_B} = \textit{Tile}(x_B[t{-}m{+}n{+}1])$ \Comment{Tile the last element of the unmasked window.} 
    \State $\mathbf{r_B} = e( [\mathbf{\Theta_B}, x_B[t{-}m{+}n{+}1:t]] \cup \mathbf{0}; \mathbf{\Omega}_e)$  \Comment{Obtain encoded representation for the unmasked input window.} 
    \State $\mathbf{b_B} = d^{(b)}(\mathbf{r_B}; \mathbf{\Omega}_{d^{(b)}})$  \Comment{Obtain backcasts for the masked window.} 
    \State $\mathbf{\Omega}_e \leftarrow \mathbf{\Omega}_e - \alpha \cdot \nabla_{\mathbf{\Omega}_e} L(\mathbf{b}, x_B[t-m+1:t])$  \Comment{Self-adaptation: Update the encoder with the backcast loss.} 
    \State $\mathbf{\Omega}_{d^{(b)}} \leftarrow \mathbf{\Omega}_{d^{(b)}} - \alpha \cdot \nabla_{\mathbf{\Omega}_{d^{(b)}}} L(\mathbf{b}, x_B[t-m+1:t)$  \Comment{Self-adaptation: Update the decoder with the backcast loss.} 
    \State $\mathbf{b_B} = d^{(b)}(\mathbf{r_B}; \mathbf{\Omega}_{d^{(b)}})$  \Comment{Obtain backcasts for the masked window.} 
    \State $\mathbf{e_B} = x_B[t-m+1:t]- \mathbf{b_B}$  \Comment{Obtain errors for the backcasts.} 
    \State $\mathbf{r_B} = e(x_B[t-m+1:t] \cup \mathbf{e_B}; \mathbf{\Omega}_e)$  \Comment{Obtain encoded representation for the input window combined with the errors.} 
    \State $\hat{y}_B = d^{(f)}(\mathbf{r_B}; \mathbf{\Omega}_{d^{(f)}})$ \Comment{Forecast.} 
    \State $\mathbf{\Omega}_e \leftarrow \mathbf{\Omega}_e - \gamma \cdot \nabla_{\mathbf{\Omega}_e} L(\hat{y}_B, y_B)$  \Comment{Update the encoder with the forecast loss.} 
    \State $\mathbf{\Omega}_{d^{(f)}} \leftarrow \mathbf{\Omega}_{d^{(f)}} - \gamma \cdot \nabla_{\mathbf{\Omega}_{d^{(f)}}} L(\hat{y}_B, y_B)$  \Comment{Update the decoder with the forecast loss.}
    \EndWhile  \label{Training loop}
\end{algorithmic}
\label{psuedocodetrain}
\end{algorithm*}

\subsection{Pseudo-code for training}

Algorithm \ref{psuedocodetrain} shows the pseudocode for training for the proposed SAF.

\subsection{Dataset and hyperparameter tuning details}

For all datasets, we use automated hyperparameter tuning based on the validation performance. For synthetic autoregressive datasets we use 100, for other datasets we use 200 trials. Tables \ref{table:hparams_syn_autoregressive} show the hyperparameter search spaces used for all datasets.

\textbf{Autoregressive processes:}
Table \ref{table:hparams_syn_autoregressive} shows the hyperparameter search space used in experiments.

\begin{table*}[!htbp]
\centering
\caption{Hyperparameter search space for synthetic autoregressive datasets.}
\begin{tabular}{|c|c|}
\cline{1-2}
\textbf{Hyperparameter} &  \textbf{Candidate values} \\ \cline{1-2}
Batch size &  [32, 64, 128, 256] \\ \cline{1-2}
Baseline learning rate & [ .0001, .0003, .001] \\ \cline{1-2}
Self-adaptation learning rate (for SAF only) & [.00003, .0001, .0003, .001] \\ \cline{1-2}
Number of units & [16, 32, 64] \\ \cline{1-2}
Number of attention heads (for TFT only) & [1, 2] \\ \cline{1-2}
Dropout keep probability (for TFT only) & [0.5, 0.8, 1.0] \\ \cline{1-2}
Encoder window length & [10, 30, 50] \\ \cline{1-2}
Representation combination & [Additive, Concatenation] \\ \cline{1-2}
Max. number of iterations & 3000 \\ \cline{1-2}
Use of backcasting errors (for SAF only) & [True, False] \\ \cline{1-2}
\cline{1-2}
\end{tabular}
\label{table:hparams_syn_autoregressive}
\end{table*}

\textbf{Covid-19:}
We adapt the data from \cite{covid_forecasting}.
The datasets consist 391 days from the start date of March 1, 2020. We use the last 14 days for testing and the preceding 14 for validation. 
Static features include `population', `income per capita', `population density per sq km', `number of households with public assistance or food stamps', `-60 population', `number of icu beds', `number of households', `number of hospitals with ratings 1-5', `air quality', `number of hospitals with non-emergency/emergency services and acute care and critical access', `hospital ratings with above/below national average'. Time-varying features include `number of deaths, cases, recovered, hospitalized, in ICU, and on ventilator', `mobility indices', `total tests', `social distancing indicators for restaurants and bars, non essential businesses, general population stay at home, school closures, mass gathering, face mask restrictions and emergency declarations', `day of the week', `max/min/mean temperature', `rainfall and snowfall', `antibody and antigen test counts', `search indices for symptoms cough, chills, anosmia, infection, chest pain, fever and shortness of breath', and `vaccination counts for one and two doses'. We employ standard normalization to all features with statistics averaged across all entities.
Table \ref{table:hparams_covid} shows the hyperparameter search space used in experiments.

\begin{table*}[!htbp]
\centering
\caption{Hyperparameter search space for Covid-19 dataset.}
\begin{tabular}{|c|c|}
\cline{1-2}
\textbf{Hyperparameter} &  \textbf{Candidate values} \\ \cline{1-2}
Batch size &  [64, 128, 256, 512] \\ \cline{1-2}
Baseline learning rate & [.0001, .0003, .001, .003] \\ \cline{1-2}
Self-adaptation learning rate (for SAF only) & [.00001, .00003, .0001, .0003, .001] \\ \cline{1-2}
Number of units & [16, 32, 64, 128, 256] \\ \cline{1-2}
Number of attention heads (for TFT only) & [1, 2, 4] \\ \cline{1-2}
Dropout keep probability (for TFT only) & [0.5, 0.8, 0.9, 1.0] \\ \cline{1-2}
Encoder window length & [10, 30, 50] \\ \cline{1-2}
Representation combination & [Additive, Concatenation] \\ \cline{1-2}
Max. number of iterations & 10000 \\ \cline{1-2}
Use of backcasting errors (for SAF only) & [True, False] \\ \cline{1-2}
\cline{1-2}
\end{tabular}
\label{table:hparams_covid}
\end{table*}

\textbf{M5:}
From the M5 dataset ~\footnote{\url{https://www.kaggle.com/c/m5-forecasting-accuracy/}}, we use the time-series for the first 200 products based on the IDs. We employ standard normalization to all features with statistics averaged across all entities.
Table \ref{table:hparams_m5} shows the hyperparameter search space used in experiments.

\begin{table*}[!htbp]
\centering
\caption{Hyperparameter search space for M5 dataset.}
\begin{tabular}{|c|c|}
\cline{1-2}
\textbf{Hyperparameter} &  \textbf{Candidate values} \\ \cline{1-2}
Batch size &  [64, 128, 256] \\ \cline{1-2}
Baseline learning rate & [.0001, .0003, .001, .003] \\ \cline{1-2}
Self-adaptation learning rate (for SAF only) & [.00003, .0001, .0003, .001] \\ \cline{1-2}
Number of units & [32, 64, 128, 256, 432] \\ \cline{1-2}
Number of attention heads (for TFT only) & [1, 2, 4, 8] \\ \cline{1-2}
Dropout keep probability (for TFT only) & [0.5, 0.8, 0.9, 1.0] \\ \cline{1-2}
Encoder window length & [20, 50, 100, 150] \\ \cline{1-2}
Representation combination & [Additive, Concatenation] \\ \cline{1-2}
Max. number of iterations & 60000 \\ \cline{1-2}
Use of backcasting errors (for SAF only) & [True, False] \\ \cline{1-2}
\cline{1-2}
\end{tabular}
\label{table:hparams_m5}
\end{table*}

\textbf{Electricity:}
We use the Electricity dataset \cite{NIPS2016_85422afb} as is. Similar to \cite{lim2019temporal}, we apply per-entity normalization, unlike other datasets. 
Table \ref{table:hparams_electricity} shows the hyperparameter search space used in experiments.

\begin{table*}[!htbp]
\centering
\caption{Hyperparameter search space for Electricity dataset.}
\begin{tabular}{|c|c|}
\cline{1-2}
\textbf{Hyperparameter} &  \textbf{Candidate values} \\ \cline{1-2}
Batch size &  [64, 128, 256, 512, 1024] \\ \cline{1-2}
Baseline learning rate & [.0001, .0003, .001, .003] \\ \cline{1-2}
Self-adaptation learning rate (for SAF only) & [.00001, .00003, .0001, .0003, .001] \\ \cline{1-2}
Number of units & [32, 64, 128, 256] \\ \cline{1-2}
Number of attention heads (for TFT only) & [1, 2, 4, 8] \\ \cline{1-2}
Dropout keep probability (for TFT only) & [0.5, 0.8, 0.9, 1.0] \\ \cline{1-2}
Encoder window length & [96, 168] \\ \cline{1-2}
Representation combination & [Additive, Concatenation] \\ \cline{1-2}
Max. number of iterations & 100000 \\ \cline{1-2}
Use of backcasting errors (for SAF only) & [True, False] \\ \cline{1-2}
\cline{1-2}
\end{tabular}
\label{table:hparams_electricity}
\end{table*}

\textbf{Crypto:}
We use the G-Research Crypto Forecasting data on Kaggle\footnote{\url{https://www.kaggle.com/c/g-research-crypto-forecasting/}}, to forecast short term returns in 14 popular cryptocurrencies. The data is since 2018. We scale the given target by 1000. We use all time-varying and static features as is. We employ standard normalization to all features with statistics averaged across all entities.

Table \ref{table:hparams_crypto} shows the hyperparameter search space used in experiments.

\begin{table*}[!htbp]
\centering
\caption{Hyperparameter search space for Crypto dataset.}
\begin{tabular}{|c|c|}
\cline{1-2}
\textbf{Hyperparameter} &  \textbf{Candidate values} \\ \cline{1-2}
Batch size &  [64, 128, 256, 512] \\ \cline{1-2}
Baseline learning rate & [.00003, .0001, .0003, .001] \\ \cline{1-2}
Self-adaptation learning rate (for SAF only) & [.00001, .00003, .0001, .0003] \\ \cline{1-2}
Number of units & [64, 128, 256] \\ \cline{1-2}
Number of attention heads (for TFT only) & [1, 2, 4] \\ \cline{1-2}
Dropout keep probability (for TFT only) & [0.5, 0.8, 1.0] \\ \cline{1-2}
Encoder window length & [60, 120, 240] \\ \cline{1-2}
Representation combination & [Additive, Concatenation] \\ \cline{1-2}
Max. number of iterations & [50000, 100000] \\ \cline{1-2}
Use of backcasting errors (for SAF only) & [True, False] \\ \cline{1-2}
\cline{1-2}
\end{tabular}
\label{table:hparams_crypto}
\end{table*}

\textbf{SP500:}
We adapt the data from Kaggle `Stock Market Data' ~\footnote{\url{https://www.kaggle.com/paultimothymooney/stock-market-data}}
From SP500 index, we choose 75 equities as the ones that have data since 01-01-2012:
`FB', `JKHY', `GOOG', `DVA', `FIS', `LBTYA', `FAST', `LNT', `FE', `ILMN', `DPZ', `ESS', `GM', `JNPR', `FFIV', `EOG', `FANG', `ICE', `DRI', `EQIX', `GWW', `EA', `DHI', `IPGP', `HES', `MKTX', `ABBV', `GILD', `EMN', `FCX', `EW', `KRA', `FN', `DG', `FDX', `FISV', `ENS', `MPC', `EXR', `LVS', `EL', `FRC', `D', `LRCX', `ISRG', `EQR', `CTQ', `EIX', `DXCM', `LNC', `CTSH', `HFC', `KEY', `IT', `HBI', `GPC', `HSY', `FBHS', `DFS', `DAL', `EFX', `GRMN', `KSS', `KMX', `FLT', `FTI', `CUK', `HRL', `ES', `DLTR', `CHTR', `EBAY', `GPN', `DRE', and `CTXS'. We use the time-varying features of `Low', `Open', `High', `Close', and `Adjusted Close' prices, and `Volume', all converted to percent change.  We employ standard normalization to all features with statistics averaged across all entities.
Table \ref{table:hparams_sp500} shows the hyperparameter search space used in experiments.

\begin{table*}[!htbp]
\centering
\caption{Hyperparameter search space for SP500 dataset.}
\begin{tabular}{|c|c|}
\cline{1-2}
\textbf{Hyperparameter} &  \textbf{Candidate values} \\ \cline{1-2}
Batch size &  [32, 64, 128, 256] \\ \cline{1-2}
Baseline learning rate & [0.0001, 0.0003, 0.001] \\ \cline{1-2}
Self-adaptation learning rate (for SAF only) & [.00001, .00003, .0001, .0003, .001] \\ \cline{1-2}
Number of units & [32, 64, 96, 128, 256] \\ \cline{1-2}
Number of attention heads (for TFT only) & [1, 2, 4, 8] \\ \cline{1-2}
Dropout keep probability (for TFT only) & [0.5, 0.8, 0.9, 1.0] \\ \cline{1-2}
Encoder window length & [96, 168] \\ \cline{1-2}
Representation combination & [Additive, Concatenation] \\ \cline{1-2}
Max. number of iterations & 40000 \\ \cline{1-2}
Use of backcasting errors (for SAF only) & [True, False] \\ \cline{1-2}
\cline{1-2}
\end{tabular}
\label{table:hparams_sp500}
\end{table*}

\newpage

\bibliography{references.bib}

\begin{thebibliography}{41}
\providecommand{\natexlab}[1]{#1}

\bibitem[{Afshin et~al.(2017)Afshin, Forouzanfar, Reitsma, Sur, Estep, Lee,
  Marczak, Mokdad, Moradi-Lakeh, Naghavi, Salama, Vos, Abate, Cristiana, Ahmed,
  Al-Aly, Alkerwi, Al-Raddadi, Amare, and Collaborators}]{obesity}
Afshin, A.; Forouzanfar, M.; Reitsma, M.; Sur, P.; Estep, K.; Lee, A.; Marczak,
  L.; Mokdad, A.; Moradi-Lakeh, M.; Naghavi, M.; Salama, J.; Vos, T.; Abate,
  K.; Cristiana, A.; Ahmed, M.; Al-Aly, Z.; Alkerwi, A.; Al-Raddadi, R.; Amare,
  A.; and Collaborators, G. 2017.
\newblock Health Effects of Overweight and Obesity in 195 Countries over 25
  Years.
\newblock \emph{New England Journal of Medicine}, 377(1): 13--27.

\bibitem[{Arik et~al.(2021)Arik, Shor, Sinha, Yoon, Ledsam, Le, Dusenberry,
  Yoder, Popendorf, Epshteyn, Euphrosine, Kanal, Jones, Li, Luan, Mckenna,
  Menon, Singh, Sun, Ravi, Zhang, Sava, Cunningham, Kayama, Tsai, Yoneoka,
  Nomura, Miyata, and Pfister}]{covid_forecasting}
Arik, S.~O.; Shor, J.; Sinha, R.; Yoon, J.; Ledsam, J.~R.; Le, L.~T.;
  Dusenberry, M.~W.; Yoder, N.~C.; Popendorf, K.; Epshteyn, A.; Euphrosine, J.;
  Kanal, E.; Jones, I.; Li, C.-L.; Luan, B.; Mckenna, J.; Menon, V.; Singh, S.;
  Sun, M.; Ravi, A.~S.; Zhang, L.; Sava, D.; Cunningham, K.; Kayama, H.; Tsai,
  T.; Yoneoka, D.; Nomura, S.; Miyata, H.; and Pfister, T. 2021.
\newblock A prospective evaluation of AI-augmented epidemiology to forecast
  {COVID-19} in the USA and Japan.
\newblock \emph{npj Digital Medicine}.

\bibitem[{Bartler et~al.(2021)Bartler, Bühler, Wiewel, Döbler, and
  Yang}]{bartler2021mt3}
Bartler, A.; Bühler, A.; Wiewel, F.; Döbler, M.; and Yang, B. 2021.
\newblock {MT3:} Meta Test-Time Training for Self-Supervised Test-Time
  Adaption.
\newblock \emph{arXiv:2103.16201}.

\bibitem[{Bayati, Khoa~Nguyen, and Cheriet(2018)}]{8490843}
Bayati, A.; Khoa~Nguyen, K.; and Cheriet, M. 2018.
\newblock Multiple-Step-Ahead Traffic Prediction in High-Speed Networks.
\newblock \emph{IEEE Communications Letters}, 22(12): 2447--2450.

\bibitem[{Box and Pierce(1970)}]{arima2}
Box, G. E.~P.; and Pierce, D.~A. 1970.
\newblock Distribution of Residual Autocorrelations in
  Autoregressive-Integrated Moving Average Time Series Models.
\newblock \emph{Journal of the American Statistical Association}, 65(332):
  1509--1526.

\bibitem[{Brahim-Belhouari and Bermak(2004)}]{BRAHIMBELHOUARI2004705}
Brahim-Belhouari, S.; and Bermak, A. 2004.
\newblock Gaussian process for nonstationary time series prediction.
\newblock \emph{Computational Statistics \& Data Analysis}, 47(4): 705--712.

\bibitem[{Cao and Gu(2002)}]{dynamic_svm}
Cao, L.; and Gu, Q. 2002.
\newblock Dynamic support vector machines for non-stationary time series
  forecasting.
\newblock \emph{Intell. Data Anal.}, 6: 67--83.

\bibitem[{Chen et~al.(2020)Chen, Li, Wu, Liang, and Jha}]{ood2}
Chen, J.; Li, Y.; Wu, X.; Liang, Y.; and Jha, S. 2020.
\newblock Robust Out-of-distribution Detection for Neural Networks.
\newblock \emph{arXiv:2003.09711}.

\bibitem[{Chen et~al.(2021)Chen, Goel, Sohoni, Poms, Fatahalian, and
  R{\'{e}}}]{Mandoline}
Chen, M.~F.; Goel, K.; Sohoni, N.~S.; Poms, F.; Fatahalian, K.; and R{\'{e}},
  C. 2021.
\newblock Mandoline: Model Evaluation under Distribution Shift.
\newblock \emph{arXiv:2107.00643}.

\bibitem[{Dahlhaus(1997)}]{Dahlhaus}
Dahlhaus, R. 1997.
\newblock {Fitting time series models to nonstationary processes}.
\newblock \emph{The Annals of Statistics}, 25(1): 1 -- 37.

\bibitem[{Fang et~al.(2020)Fang, Lu, Niu, and Sugiyama}]{Fang_rethinking}
Fang, T.; Lu, N.; Niu, G.; and Sugiyama, M. 2020.
\newblock Rethinking Importance Weighting for Deep Learning under Distribution
  Shift.
\newblock \emph{arXiv:2006.04662}.

\bibitem[{Finn, Abbeel, and Levine(2017)}]{maml}
Finn, C.; Abbeel, P.; and Levine, S. 2017.
\newblock Model-Agnostic Meta-Learning for Fast Adaptation of Deep Networks.
\newblock \emph{arXiv:1703.03400}.

\bibitem[{Ganin et~al.(2016)Ganin, Ustinova, Ajakan, Germain, Larochelle,
  Laviolette, Marchand, and Lempitsky}]{ganin2016domainadversarial}
Ganin, Y.; Ustinova, E.; Ajakan, H.; Germain, P.; Larochelle, H.; Laviolette,
  F.; Marchand, M.; and Lempitsky, V. 2016.
\newblock Domain-Adversarial Training of Neural Networks.
\newblock \emph{arXiv:1505.07818}.

\bibitem[{Gardner~Jr(1985)}]{gardner1985exponential}
Gardner~Jr, E.~S. 1985.
\newblock Exponential smoothing: The state of the art.
\newblock \emph{Journal of forecasting}, 4(1): 1--28.

\bibitem[{Gfeller et~al.(2020)Gfeller, Frank, Roblek, Sharifi, Tagliasacchi,
  and Velimirovi{\'c}}]{selfsupverised_pitch}
Gfeller, B.; Frank, C.; Roblek, D.; Sharifi, M.; Tagliasacchi, M.; and
  Velimirovi{\'c}, M. 2020.
\newblock SPICE: Self-supervised pitch estimation.
\newblock \emph{IEEE Trans Audio, Speech, and Language Processing}, 28:
  1118--1128.

\bibitem[{Glantz and Kissell(2013)}]{multi_asset_risk}
Glantz, M.; and Kissell, R. 2013.
\newblock Multi-Asset Risk Modeling: Techniques for a Global Economy in an
  Electronic and Algorithmic Trading Era.
\newblock \emph{Multi-Asset Risk Modeling: Techniques for a Global Economy in
  an Electronic and Algorithmic Trading Era}, 1--516.

\bibitem[{Goodfellow et~al.(2016)Goodfellow, Bengio, Courville, and
  Bengio}]{goodfellow2016deep}
Goodfellow, I.; Bengio, Y.; Courville, A.; and Bengio, Y. 2016.
\newblock \emph{Deep learning}, volume~1.
\newblock MIT Press.

\bibitem[{Hansen et~al.(2021)Hansen, Jangir, Sun, Alenyà, Abbeel, Efros,
  Pinto, and Wang}]{hansen2021selfsupervised}
Hansen, N.; Jangir, R.; Sun, Y.; Alenyà, G.; Abbeel, P.; Efros, A.~A.; Pinto,
  L.; and Wang, X. 2021.
\newblock Self-Supervised Policy Adaptation during Deployment.
\newblock \emph{arXiv:2007.04309}.

\bibitem[{Harvey(1990)}]{harvey1990forecasting}
Harvey, A.~C. 1990.
\newblock \emph{Forecasting, Structural Time Series Models and the Kalman
  Filter}.
\newblock Cambridge University Press.

\bibitem[{Hendrycks and Gimpel(2016)}]{ood1}
Hendrycks, D.; and Gimpel, K. 2016.
\newblock A Baseline for Detecting Misclassified and Out-of-Distribution
  Examples in Neural Networks.
\newblock \emph{arXiv:1610.02136}.

\bibitem[{Hillmer and Tiao(1982)}]{arima1}
Hillmer, S.~C.; and Tiao, G.~C. 1982.
\newblock An ARIMA-Model-Based Approach to Seasonal Adjustment.
\newblock \emph{Journal of the American Statistical Association}, 77(377):
  63--70.

\bibitem[{Kang et~al.(2019)Kang, Jiang, Yang, and
  Hauptmann}]{Contrastive_adaptation}
Kang, G.; Jiang, L.; Yang, Y.; and Hauptmann, A.~G. 2019.
\newblock Contrastive Adaptation Network for Unsupervised Domain Adaptation.
\newblock \emph{arXiv:1901.00976}.

\bibitem[{Koh et~al.(2020)Koh, Sagawa, Marklund, Xie, Zhang, Balsubramani, Hu,
  Yasunaga, Phillips, Beery, Leskovec, Kundaje, Pierson, Levine, Finn, and
  Liang}]{wilds}
Koh, P.~W.; Sagawa, S.; Marklund, H.; Xie, S.~M.; Zhang, M.; Balsubramani, A.;
  Hu, W.; Yasunaga, M.; Phillips, R.~L.; Beery, S.; Leskovec, J.; Kundaje, A.;
  Pierson, E.; Levine, S.; Finn, C.; and Liang, P. 2020.
\newblock {WILDS:} {A} Benchmark of in-the-Wild Distribution Shifts.
\newblock \emph{arXiv:2012.07421}.

\bibitem[{Kuznetsov and Mohri(2015)}]{NIPS2015_41f1f191}
Kuznetsov, V.; and Mohri, M. 2015.
\newblock Learning Theory and Algorithms for Forecasting Non-stationary Time
  Series.
\newblock In \emph{NIPS}.

\bibitem[{Kuznetsov and Mohri(2020)}]{Kuznetsov2020}
Kuznetsov, V.; and Mohri, M. 2020.
\newblock Discrepancy-Based Theory and Algorithms for Forecasting
  Non-Stationary Time Series.
\newblock \emph{Annals of Mathematics and Artificial Intelligence}, 88(4):
  367--399.

\bibitem[{Lim et~al.(2021)Lim, Arik, Loeff, and Pfister}]{lim2019temporal}
Lim, B.; Arik, S.~O.; Loeff, N.; and Pfister, T. 2021.
\newblock Temporal Fusion Transformers for Interpretable Multi-horizon Time
  Series Forecasting.
\newblock \emph{International Journal of Forecasting}, 37(4): 1748--1764.

\bibitem[{Lindemann et~al.(2021)Lindemann, Müller, Vietz, Jazdi, and
  Weyrich}]{LINDEMANN2021650}
Lindemann, B.; Müller, T.; Vietz, H.; Jazdi, N.; and Weyrich, M. 2021.
\newblock A survey on long short-term memory networks for time series
  prediction.
\newblock \emph{Procedia CIRP}, 99: 650--655.
\newblock 14th CIRP Conference on Intelligent Computation in Manufacturing
  Engineering, 15-17 July 2020.

\bibitem[{Matic, Packham, and Härdle(2021)}]{matic2021hedging}
Matic, J.~L.; Packham, N.; and Härdle, W.~K. 2021.
\newblock Hedging Cryptocurrency Options.
\newblock arXiv:2112.06807.

\bibitem[{Montero{-}Manso and Hyndman(2020)}]{Montero2020}
Montero{-}Manso, P.; and Hyndman, R.~J. 2020.
\newblock Principles and Algorithms for Forecasting Groups of Time Series:
  Locality and Globality.
\newblock \emph{CoRR}, abs/2008.00444.

\bibitem[{Ng and Young(1990)}]{recursive_estimation_nonstationary}
Ng, C.~N.; and Young, P.~C. 1990.
\newblock Recursive estimation and forecasting of non-stationary time series.
\newblock \emph{Journal of Forecasting}, 9(2): 173--204.

\bibitem[{Oreshkin et~al.(2019)Oreshkin, Carpov, Chapados, and
  Bengio}]{oreshkin2019nbeats}
Oreshkin, B.~N.; Carpov, D.; Chapados, N.; and Bengio, Y. 2019.
\newblock N-BEATS: Neural basis expansion analysis for interpretable time
  series forecasting.
\newblock \emph{arXiv:1905.10437}.

\bibitem[{Ovadia et~al.(2019)Ovadia, Fertig, Ren, Nado, Sculley, Nowozin,
  Dillon, Lakshminarayanan, and Snoek}]{ovadia2019trust}
Ovadia, Y.; Fertig, E.; Ren, J.; Nado, Z.; Sculley, D.; Nowozin, S.; Dillon,
  J.~V.; Lakshminarayanan, B.; and Snoek, J. 2019.
\newblock Can You Trust Your Model's Uncertainty? Evaluating Predictive
  Uncertainty Under Dataset Shift.
\newblock \emph{arXiv:1906.02530}.

\bibitem[{Rhif et~al.(2019)Rhif, Ben~Abbes, Farah, Martínez, and
  Sang}]{app9071345}
Rhif, M.; Ben~Abbes, A.; Farah, I.~R.; Martínez, B.; and Sang, Y. 2019.
\newblock Wavelet Transform Application for/in Non-Stationary Time-Series
  Analysis: A Review.
\newblock \emph{Applied Sciences}, 9(7).

\bibitem[{Salinas et~al.(2020)Salinas, Flunkert, Gasthaus, and
  Januschowski}]{deepar}
Salinas, D.; Flunkert, V.; Gasthaus, J.; and Januschowski, T. 2020.
\newblock DeepAR: Probabilistic forecasting with autoregressive recurrent
  networks.
\newblock \emph{International Journal of Forecasting}, 36(3): 1181--1191.

\bibitem[{Schneider et~al.(2020)Schneider, Rusak, Eck, Bringmann, Brendel, and
  Bethge}]{covariate_shift}
Schneider, S.; Rusak, E.; Eck, L.; Bringmann, O.; Brendel, W.; and Bethge, M.
  2020.
\newblock Improving robustness against common corruptions by covariate shift
  adaptation.
\newblock In Larochelle, H.; Ranzato, M.; Hadsell, R.; Balcan, M.; and Lin, H.,
  eds., \emph{Advances in Neural Information Processing Systems}, volume~33,
  11539--11551. Curran Associates, Inc.

\bibitem[{Sproles(1981)}]{retail}
Sproles, G.~B. 1981.
\newblock Analyzing Fashion Life Cycles—Principles and Perspectives.
\newblock \emph{Journal of Marketing}, 45(4): 116--124.

\bibitem[{Sun et~al.(2019)Sun, Wang, Liu, Miller, Efros, and Hardt}]{ttt}
Sun, Y.; Wang, X.; Liu, Z.; Miller, J.; Efros, A.~A.; and Hardt, M. 2019.
\newblock Test-Time Training for Out-of-Distribution Generalization.
\newblock \emph{arXiv:1909.13231}.

\bibitem[{Sutskever, Vinyals, and Le(2014)}]{sutskever2014sequence}
Sutskever, I.; Vinyals, O.; and Le, Q.~V. 2014.
\newblock Sequence to Sequence Learning with Neural Networks.
\newblock \emph{arXiv:1409.3215}.

\bibitem[{Wen et~al.(2017)Wen, Torkkola, Narayanaswamy, and Madeka}]{MQRNN}
Wen, R.; Torkkola, K.; Narayanaswamy, B.; and Madeka, D. 2017.
\newblock A Multi-Horizon Quantile Recurrent Forecaster.
\newblock In \emph{NIPS Workshops}.

\bibitem[{Yu, Rao, and Dhillon(2016)}]{NIPS2016_85422afb}
Yu, H.-F.; Rao, N.; and Dhillon, I.~S. 2016.
\newblock Temporal Regularized Matrix Factorization for High-dimensional Time
  Series Prediction.
\newblock In \emph{NIPS}.

\bibitem[{Zhou et~al.(2021)Zhou, Zhang, Peng, Zhang, Li, Xiong, and
  Zhang}]{zhou2021informer}
Zhou, H.; Zhang, S.; Peng, J.; Zhang, S.; Li, J.; Xiong, H.; and Zhang, W.
  2021.
\newblock Informer: Beyond Efficient Transformer for Long Sequence Time-Series
  Forecasting.
\newblock \emph{arXiv:2012.07436}.

\end{thebibliography}

\end{document}